\documentclass[runningheads,a4paper]{llncs}
\usepackage{rotating} 
\usepackage{amssymb}
\usepackage{graphicx}
\usepackage{subfig}
\usepackage{url}

\urldef{\mailsa}\path|{jdmitrie, fverbeek}@liacs.nl|
\newcommand{\keywords}[1]{\par\addvspace\baselineskip
\noindent\keywordname\enspace\ignorespaces#1}

\begin{document}

\mainmatter  

\title{Creating a new Ontology: a Modular Approach}

\titlerunning{Creating a new Ontology: a Modular Approach}

%
%
\author{Julia Dmitrieva \and Fons J.~Verbeek}
\authorrunning{Creating a new Ontology: a Modular Approach}

\institute{Universiteit Leiden, Leiden Institute of Advanced Computer Science,\\
Niels Bohrweg 1,
2333 CA  Leiden,
The Netherlands\\
\mailsa\\
\url{http://bio-imaging.liacs.nl/}}

%
%

\maketitle


\section{Introduction}
Ontologies in life sciences, in particular, members of the \textsc{OBO Foundry}~\cite{OBO},
contain information about species, proteins, chemicals, genomes, pathways, diseases, etc.
Information in these ontologies might 
overlap, and it is possible that a certain concept is defined in different
ontologies from a different point of view and at different level of granularity.
Therefore, the combination of information from different ontologies
is useful to create a new ontology.
\paragraph{\textbf{Case Study}}\label{ModuleIntegraion:example}
The integration will be illustrated with a case study on Toll-like receptors.
If we want to investigate what kind of information about Toll-like receptors
is available in \textsc{Molecule Role Ontology} (MoleculeRoleOntlogy)~\cite{OBO},
then we will see that Toll-like receptors are defined as pattern recognition receptors.
In the \textsc{Biological Process Ontology} (GO)~\cite{OBO} the Toll-like receptors
are described in the context of signaling pathway and are subsumed by the  \emph{pattern recognition receptor signaling pathway}.
In the \textsc{Protein}~\cite{OBO} ontology a Toll-like receptor is just a protein.
In the \textsc{NCI\_Thesaurus}~\cite{NCIThesaurus} ontology Toll-like receptors are defined 
as \emph{Cell Surface Receptors}.
It follows from foregoing that multiple ontologies model different aspects of 
the same concept and the combination of the available information provides
more knowledge about concepts where an ontology developer is interested in.

We introduce an approach for generating a new ontology in which 
ontologies from \textsc{OBO Foundry} are reused. First, we extract modules from these ontologies, on the basis of the  well defined
modularity approach~\cite{DBLP:series/lncs/GrauHKS09}.
As a signature for the modules we are using the symbols that match the 
terms of interest as indicated by the user. In our case study we create an ontology 
about Toll-like receptors, therefore we use two \emph{seed} terms (Toll, TLR).
Subsequently, we create mappings between concepts in the modules. It has already been shown~\cite{SimpleMethods} 
that the simple similarity algorithms outperform structural similarity algorithms in biomedical ontologies. 
To this end, we have based our mappings on the similarity distance~\cite{Levenshtein} 
between labels and synonyms of classes in the modules. Finally, a new ontology is
created where the mappings are represented by means of \textsc{owl:equivalentClass} axiom
and small concise modules are imported. 

\section{Modules from Enriched Signature}
In our case study we have used the following biomedical ontologies obtained 
from \textsc{OBO Foundry}:
National Cancer Institute Ontology (\textsc{NCI\_Thesaurus}), GO Ontology (\textsc{GO}),
Protein Ontology (\textsc{PRO}),
Dendritic Cell ontology\\
(\textsc{dendritic\_cell}), Pathway ontology (\textsc{pathway}), Molecule Role Ontology\\
 (\textsc{MoleculeRoleOntology}), Gene Regulation Ontology (\textsc{gene\_regulation}),
and finally, Medical Subject Heading ontology (\textsc{MeSH}). All of these ontologies are in OBO format,
except for the \textsc{NCI\_Thesaurus}, which is in OWL format. 

A module comprises knowledge of a part of the domain that is dedicated
to a set of terms of user interest (\emph{seed terms}). Let  $T_1 = \{ Toll, TLR\}$ be this set.
Let $S_1$ be a set of terms (signature) from the ontology $O_1$ that represents the classes whose labels,
descriptions, ID, or other annotation properties contain the symbols from $T_1$.
The first module that we have extracted is the module from \textsc{NCI\_Thesaurus} $M_1$.
This is chosen because it is the largest ontology containing the most matches.
In order to generate a signature for the next ontology $O_2$, we are using not only
the terms from $T_1$ but we enrich this set with the terms from the module $M_1$.
The same procedure is applied
for the rest of the ontologies, namely module $M_i$ is extracted on the basis
of the terms $T_{i} = Sig(M_{i-1}) \cup T_{i-1}$. 
This method has two drawbacks. First, it depends on the order of
ontologies. 
Second, with the generation of the  new module $M_i$ new
symbols can be introduced that will match symbols from ontologies used 
in previous steps. These problems can be solved with the generation of a fixpoint.

\paragraph{\textbf{Fixpoint Modules}} \label{ModuleIntegration:fixpoint}
We have investigated whether or not we will find a fixpoint with our
module extraction method. The fixpoint is reached at the moment the
set of terms $T$ which is used in order to generate  modules during step $t_i$ does not change any more after another run with all ontologies.
This can be written as  $\cup_{k=1}^{n} Sig(M_{k,i}) = \cup_{k=1}^{n} Sig(M_{k,i+1})$, where $M_{k,i}$ is
the module $k$ created during step $t_i$. It can be formulated in a ''fixpoint-like'' way $Match(T) = T$. 

The fixpoint was reached with the following sizes of the modules, see Table~\ref{fixpoinMod}.
\begin{table}
\vspace{-15pt}
\caption{The size of the modules after  reaching the fixpoint}
\label{fixpoinMod}
 \begin{center}
\begin{tabular}{ll}
 \hline\noalign{\smallskip}
module & size in KB \\
\noalign{\smallskip}
\hline
\noalign{\smallskip}
Toll\_from\_gene\_regulation & 88.7\\
Toll\_from\_protein & 23.4 \\
Toll\_from\_chebi & 218.6 \\
Toll\_from\_mesh & 59.2 \\
Toll\_from\_dendritic\_cell & 4.2 \\
Toll\_from\_pathway & 4.1 \\
Toll\_from\_cellular\_component & 35.4 \\
Toll\_from\_molecular\_function & 11.4 \\
Toll\_from\_MoleculeRoleOntology & 46.9 \\
Toll\_from\_biological\_process & 221.1 \\
Toll\_from\_Thesaurus & 802.1 \\
\hline
\end{tabular}
\end{center}
\vspace{-15pt}
\end{table}
\section{Ontology Mapping}
In this paper we use a more loosely definition of the concept \emph{mapping}
compared with the definition given in \cite{DBLP:conf/dagstuhl/KalfoglouS05}
in which mapping is a morphism.
In our approach \emph{mapping} is a partial function that maps from 
subset $S_1 \subseteq Sig(O_1)$ to subset $S_2 \subseteq Sig(O_2)$.
We deliberatively reject the morphism requirement, thus, the structural
dependencies will not be preserved after mapping, because we are 
interested in consequents of this mapping to the original ontologies,
namely, whether and how the structural dependences will be broken.

For our experimental prototype system we use our own mappings based on
the syntactic similarity.
It has been already shown~\cite{SimpleMethods} that in the case of biomedical ontologies
the simple mappings methods are sufficient and outperform more complex methods.

We compare  characteristics (id, label, description) for all classes from ontology $O_1$
with the same characteristics for all classes from ontology $O_2$.
The comparison is  based on the Levenshtein distance algorithm \cite{Levenshtein}.
We have adapted the Levenshtein distance and introduce a metric $Lev$ (in the range $[0 \ldots 1]$).
Two classes $C_i$ and $C_j$ are considered to be similar 
if they have the maximum value for $Lev$ metric and if this value is also higher than the threshold $t=0.95$
that was experimentally determined.

\section{Integration Information from Ontologies}
The final step of the ontology creation is the integration of the modules
into one ontology.
If there a mapping exists between two classes $C_i$ and $C_j$
from the modules $M_i$ and $M_j$ respectively we add the equivalence relation\\
\textsc{owl:equivalentClass} between these classes in the new ontology.
Besides the equivalence relationships  the new ontology contains the
\textsc{OWL:imports} axioms, where all the  created modules are imported.

So far, this all seems rather straightforward. However,
the problem with this integrated ontology $O_{1 \ldots n}$ is that 
it contains many unsatisfiable classes. In order to understand the
reason of this unsatisfiability we have applied different experiments.
First, we have merged all pairs of the modules, namely $\forall_{i\not = j} O_{i,j} \equiv M_i \cup M_j $.
For each merged ontology $O_{i,j}$ we have checked for unsatisfiable classes.
Already at this stage of integration different merged pairs contain
unsatisfiable classes. We have used the Pellet~\cite{DBLP:journals/ws/SirinPGKK07} reasoner in order to reveal
the explanations of unsatisfiability. After we have repaired unsatisfiable classes in the merged 
pairs of ontologies $O_{i,j}$ we have had to check satisfiability of the integrated ontology $O_{1 \ldots n}$. 
There were still $46$ unsatisfiable classes. The unsatisfiabilities in the integrated ontology
have also been solved by means of Pellet reasoner explanations.

\section{Conclusion}
We have described a method to generate a new ontology on the basis of the 
bio-ontologies most of which are available in \textsc{OBO Foundry}.
We have shown how to create modules on the basis of the terms of
interest. The signature for the  module extraction is enriched
by the symbols from other modules with the fixpoint as a stop criterion.
We have integrated modules on the basis of mappings created using Levenshtein
distance similarity.

We have investigated how to solve unsatisfiable classes which
appear after the integration of the modules. Although the number of 
unsatisfiable classes was high, it was possible to solve unsatisfiabilities
with the help of explanations provided by the Pellet reasoner.

In this study we have shown that the  modularity and simple mappings
provide a good foundation for the creation of a new ontology in an pseudo-automated way.
This method can be used when
an ontology engineer does not want to create a new ontology from scratch, but rather wants to
reuse knowledge already presented in other ontologies. Moreover, this is the strategy
that should be preferred and has to be applied more often as ontologies gain importance
in life sciences.
\bibliographystyle{splncs03}
\bibliography{bibliography}
\end{document}